\documentclass[sigconf]{acmart}
\setcopyright{none}
\renewcommand\footnotetextcopyrightpermission[1]{}
\acmConference[ICAIF '26]{7th ACM International Conference on AI in Finance}{2026}{Milan, Italy}
\usepackage{booktabs}
\usepackage{tikz}
\usetikzlibrary{arrows.meta,positioning,fit,backgrounds,calc,shapes.geometric}
\definecolor{acc}{RGB}{31,78,121}
\definecolor{hl}{RGB}{176,58,46}
\definecolor{amb}{RGB}{193,145,40}
\definecolor{grn}{RGB}{46,111,64}
\definecolor{nul}{RGB}{122,104,166}
\definecolor{spp}{RGB}{84,71,140}
\usepackage{enumitem}
\newcommand{\car}{\mathrm{CAR}[-1,+1]}
\makeatletter
\g@addto@macro\normalsize{%
  \setlength\abovedisplayskip{5pt plus 1pt minus 2pt}%
  \setlength\belowdisplayskip{5pt plus 1pt minus 2pt}%
  \setlength\abovedisplayshortskip{3pt plus 1pt}%
  \setlength\belowdisplayshortskip{3pt plus 1pt}}
\makeatother

\begin{document}

\title[Agent Policy-Value Audit]{Agent Policy-Value Audit: Separating Transition Composition from Event Selection in Financial LLM Agents}

\author{Mingyang (Alex) Chen}\affiliation{\institution{Emory University}\city{Atlanta}\country{USA}}
\author{Yida (Andrew) Xu}\affiliation{\institution{Emory University}\city{Atlanta}\country{USA}}
\author{Huiwen (Aurora) Chen}\affiliation{\institution{Emory University}\city{Atlanta}\country{USA}}
\author{Yiming Lu}\affiliation{\institution{Emory University}\city{Atlanta}\country{USA}}
\author{Wei Jin}\affiliation{\institution{Emory University}\city{Atlanta}\country{USA}}

\begin{abstract}
Financial LLM agents are often evaluated by comparing their end-to-end returns with those of a baseline and testing the paired difference against zero. This measures whether deploying the agent changes realized performance, but it does not isolate event-selection skill. An agent that frequently changes positions from flat to long can earn a positive paired return from an upward-drifting event pool even when it selects events at random. We propose the Agent Policy-Value Audit, which holds fixed the observed count of each ordered action-change type and randomly reassigns them across eligible events. The average payoff from these reassignments is the composition benchmark; the difference between observed deployment value and this benchmark is selection value. In semi-synthetic benchmarks based on real earnings-event returns, a zero-centered paired test falsely attributes passive exposure to selection skill in $11.6\%$ of no-skill replications, while the transition-matched audit reduces this rate to $5.3\%$. Applied retrospectively to 723 earnings events at 44 U.S. consumer-facing firms, the audit decomposes the agent's gross deployment value of $+15.2$ bps/event into a $+25.8$ composition benchmark and a $-10.6$ selection value. The agent does not detectably outperform matched random assignments. Financial-agent evaluations should report deployment value separately from event-selection value.
\end{abstract}

\ccsdesc[500]{Applied computing~Economics}
\ccsdesc[300]{Computing methodologies~Machine learning}

\keywords{LLM agents; policy evaluation; audit; evaluator benchmark; event-driven trading; backtest integrity; clustered inference}

\maketitle

\section{Introduction}\label{sec:intro}
Large language model (LLM) agents are being built into financial analysis and decision pipelines: multi-agent systems assign analyst, trader, and risk-manager roles \citep{tradingagents}, ingest layered agent memory  \citep{yu2024finmem}, and build on open financial backbones \citep{yang2023fingpt}. Before such agents are trusted with capital, an evaluation should establish that reported gains come from the agent's decisions rather than from the market environment it trades in.

Most existing financial-agent evaluations primarily answer a deployment question: \emph{does deploying the assembled system improve realized performance relative to a baseline?} End-to-end backtests report system returns; deflated performance metrics and multiple-testing corrections address strategy search; and walk-forward designs address temporal validation \citep{bailey2014dsr,bailey2014charlatanism,harvey2016,finsaber}. These tools are important, but they generally do not isolate a separate event-selection question: holding fixed the mix of action changes an agent makes, were those changes assigned to unusually favorable events?

Pairing the agent's actions with the baseline and testing the return difference against zero correctly measures incremental deployment value. For event selection, however, zero is generally not the right reference point. Consider an agent that frequently changes positions from flat to long. Even if those changes are assigned to events at random, they can have a positive expected payoff when the eligible event pool has a positive average return. More generally, each transition inherits the average return of its eligible event pool, so the agent's realized transition composition can have nonzero expected payoff without any event-selection skill. A zero-centered test may therefore mistake the payoff of the agent's trading style for evidence that it selected the right events. The central question is what benchmark should replace zero when evaluating event selection.

To address this question, we propose the \emph{Agent Policy-Value Audit}, a benchmark constructed from the agent's own trades. The audit scores the agent and the baseline on the same events, with agreement events contributing zero; holds fixed the observed number and types of action changes; and reassigns those changes at random across eligible events. The average payoff of these reassignments is the \emph{composition benchmark}: what the agent's transition composition earns without any event selection. Deployment value then splits into two parts---deployment value $=$ composition benchmark $+$ selection value---and only the selection value reflects which events received the changes. Two safeguards sit outside this decomposition: confidence intervals let returns of the same firm move together (firm-clustered inference), and when part of the scored return is already observable at decision time, only the future part is scored (future-only scoring). The paper makes two contributions:
\begin{itemize}[leftmargin=1.2em,itemsep=1pt,topsep=2pt,parsep=0pt]
\item \textbf{A transition-matched audit for two distinct evaluation questions.} The audit evaluates incremental deployment value against zero, but evaluates event selection against a reassignment benchmark that preserves the agent's observed transition composition. This prevents the mechanical payoff of the agent's trading style from being attributed to event-selection skill.
\item \textbf{A controlled evaluator benchmark and an empirical application.} Semi-synthetic tests show when conventional paired evaluations misattribute passive exposure to event selection, and the retrospective application shows that a positive deployment estimate and a negative selection estimate can coexist.
\end{itemize}

Across $5{,}000$ no-skill replications preserving the agent's observed mix of action changes, the false-attribution rate falls from $11.6\%$ under a zero-centered paired test to $5.3\%$ under the matched audit, near the $5\%$ nominal level (\S\ref{sec:poscontrol}). Applied retrospectively to frozen traces covering $723$ earnings events at $44$ U.S.\ consumer-facing firms, the audit decomposes $+15.2$ bps/event of gross deployment value into a $+25.8$ composition benchmark and a $-10.6$ selection value, indicating trading-style exposure rather than detectable event-selection skill
(\S\ref{sec:results}).

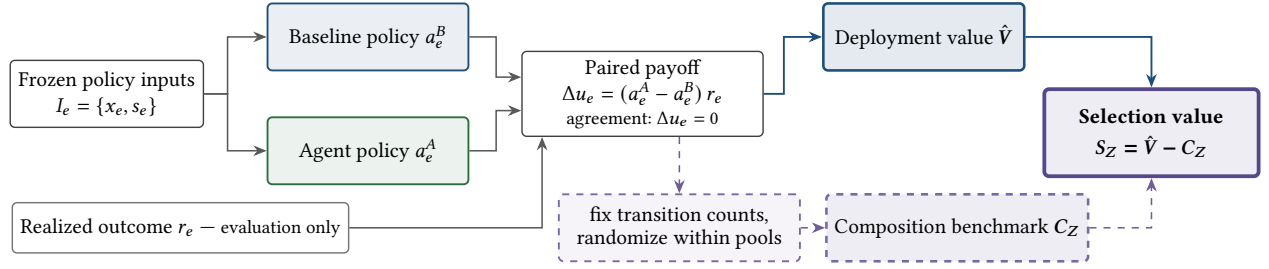
\begin{figure*}[t]
\centering
\resizebox{0.93\textwidth}{!}{%
\begin{tikzpicture}[
  font=\normalsize,
  >={Stealth[length=2mm,width=1.6mm]},
  box/.style   ={draw=black!75, line width=0.6pt, rounded corners=2pt,
                 align=center, minimum height=10mm, inner sep=3.5pt, fill=white},
  polb/.style  ={box, draw=acc, line width=0.8pt, fill=acc!9,  minimum width=30mm},
  pola/.style  ={box, draw=grn, line width=0.8pt, fill=grn!9,  minimum width=30mm},
  vbox/.style  ={box, draw=acc, line width=1.1pt, fill=acc!12, minimum width=30mm},
  nmech/.style ={box, draw=nul, line width=0.8pt, dashed, fill=nul!7, minimum width=36mm},
  cbox/.style  ={box, draw=nul, line width=1.1pt, dashed, fill=nul!12, minimum width=32mm},
  sbox/.style  ={box, draw=spp, line width=1.6pt, fill=spp!10, minimum width=32mm, minimum height=13mm},
  strip/.style ={draw=black!45, line width=0.5pt, rounded corners=1.8pt, fill=black!3,
                 font=\normalsize, inner sep=3pt},
  edge/.style  ={->, draw=black!65, line width=0.65pt},
  nedge/.style ={->, draw=nul!90!black, line width=0.65pt, dashed},
  vedge/.style ={->, draw=acc, line width=0.8pt}
]
\node[box, minimum width=26mm] (in) at (0,0) {Frozen policy inputs\\$I_e=\{x_e,s_e\}$};
\node[box, minimum width=30mm, draw=black!55, minimum height=7.5mm] (out) at (1.1,-2.0)
  {Realized outcome $r_e$ --- {\small evaluation only}};
\node[polb] (pb) at (3.9, 0.85) {Baseline policy $a^{B}_{e}$};
\node[pola] (pa) at (3.9,-0.85) {Agent policy $a^{A}_{e}$};
\node[box, minimum width=36mm] (du) at (8.0,0)
  {Paired payoff\\$\Delta u_e=(a^{A}_{e}-a^{B}_{e})\,r_e$\\[-1pt]{\small agreement: $\Delta u_e=0$}};
\node[vbox] (V) at (12.2, 0.85) {Deployment value $\boldsymbol{\hat V}$};
\node[nmech] (mech) at (8.55,-2.0) {fix transition counts,\\[-1pt]randomize within pools};
\node[cbox] (C) at (12.7,-2.0) {Composition benchmark $\boldsymbol{C_Z}$};
\node[sbox] (S) at (15.6,-0.58) {\textbf{Selection value}\\[2pt]$\boldsymbol{S_Z=\hat V-C_Z}$};
\draw[edge] (in.east) -- ++(0.35,0) |- (pb.west);
\draw[edge] (in.east) -- ++(0.35,0) |- (pa.west);
\draw[edge] (pb.east) -- ++(0.4,0) |- ($(du.west)+(0,0.25)$);
\draw[edge] (pa.east) -- ++(0.4,0) |- ($(du.west)+(0,-0.25)$);
\draw[edge] (out.east) -| ($(du.south)+(-1.5,0)$);
\draw[vedge] (du.east) -- ++(0.3,0) |- (V.west);
\draw[nedge] ($(du.south)+(0.55,0)$) -- (mech.north);
\draw[nedge] (mech.east) -- (C.west);
\draw[vedge] (V.east) -| (S.north);
\draw[nedge] (C.east) -| (S.south);
\end{tikzpicture}%
}
\Description{Flow diagram: both policies read the same frozen inputs; paired payoffs give the deployment value, which is compared with a composition benchmark built by reassigning the observed action changes within pools; the difference is the selection value.}
\caption{\textbf{Agent Policy-Value Audit.} Both policies act on the same frozen inputs, and realized outcomes enter evaluation only. The observed deployment value $\hat V$ is compared with a benchmark $C_Z$ that preserves the agent's transition composition while reassigning eligible events; the difference $S_Z$ is the selection value. Safeguards: firm-clustered uncertainty on $\hat V$; future-only scoring under decision-time overlap; leakage diagnostics on inputs.}
\label{fig:protocol}
\end{figure*}

\section{Related Work}\label{sec:related}
Financial LLM agents such as TradingAgents and FinMem \citep{tradingagents,yu2024finmem} are typically evaluated by backtesting the assembled system or by standalone return prediction, with mixed reported performance \citep{finsaber,lopezlira2023chatgpt}; recent ICAIF work evaluates agents through agent-as-a-judge benchmarks, behavioral-bias audits, and structured workflows \citep{finresearchbench2025,finethics2025,tsagent2025}---evaluations of system-level performance and output quality that do not condition attribution on the realized composition and timing of the agent's action transitions.

The closest methods come from policy evaluation and randomization inference. Off-policy evaluation solves a missing-data problem: rewards are observed only for the actions actually taken, so valuing another policy requires propensity or outcome models \citep{dudik2011doubly,swaminathan2015counterfactual,jiang2016doubly}. Here, with negligible market impact, the realized announcement return prices every candidate position, and both policies can be scored exactly on the same events. Randomization inference provides exact comparisons under a specified assignment mechanism \citep{fisher1935design,rosenbaum2002observational}, but a conventional zero-centered paired statistic does not preserve the agent's realized transition composition, so its no-selection center differs from zero whenever that composition has nonzero expected payoff. Preserving the realized transition counts within eligible baseline-action pools changes what the test measures, not the test mechanics: each intervention is compared with random interventions of the same type, so the correct center is the expected payoff of the agent's own composition (\S\ref{sec:closedform}). Because assignment is not physically randomized, the $p$-value is valid under an explicit exchangeability assumption (\S\ref{sec:inference}) rather than by design. Our contribution is not a new permutation algorithm, but an evaluation target and audit protocol: conditioning on the realized transition composition makes the reference distribution answer event selection rather than deployment, and the accompanying evaluator benchmark shows which components correct exposure misattribution and which correct decision-time overlap.

The composition/selection split echoes characteristic-matched performance attribution, which credits a manager only with returns beyond a matched portfolio \citep{daniel1997measuring}; the audit instead matches on the realized action transitions of a policy pair and supplies an exact randomization reference for the selection term. The empirical design draws on the event-study literature \citep{bernardthomas1989,mackinlay1997event}, on evidence that earnings components and expectation errors predict subsequent returns \citep{skinnersloan2002,doyle2003predictive}, and on investor-attention and social-media research \citep{da2011attention,chen2014wisdom,cookson2020disagree}. Backtest-integrity tools address overfitting and multiple testing \citep{bailey2014dsr,bailey2014charlatanism,harvey2016,romanowolf2005}, alongside preregistration practices \citep{nosek2018preregistration}; the checks in \S\ref{sec:protocol} and \S\ref{sec:results} add exposure attribution and decision-time overlap to that list.

\section{The Agent Policy-Value Audit}\label{sec:protocol}

The audit evaluates an event-aligned trading policy against a prespecified baseline. Each event $e$ is a quarterly earnings announcement with a realized three-day risk-adjusted return $r_e$, measured in basis points (bps) as a cumulative abnormal return (construction in \S\ref{sec:data}). Both policies observe the same frozen pre-decision information and take a long, short, or flat position; the paired payoff of an event is the position difference multiplied by $r_e$, so events on which the two policies agree contribute zero. We call an event on which the policies disagree an \emph{intervention}, and its \emph{transition} is the ordered action change (flat to long, for example). Figure~\ref{fig:protocol} shows the flow.

\subsection{Evaluation protocol}\label{sec:totalvalue}
The first quantity answers the deployment question. With $a^{A}_e$ the agent's position and $a^{B}_e$ the baseline's, the paired payoff and the \emph{deployment value} are
\begin{equation}\label{eq:deployment}
\Delta u_e=(a^{A}_e-a^{B}_e)\,r_e,\qquad
\hat V=\frac{1}{N}\sum_{e}\Delta u_e .
\end{equation}
Here $N$ counts all scored events, including agreements, which contribute zero; $\hat V$ is reported per event so that policies with different trading rates remain comparable.

\label{sec:benchmarkdef}\label{sec:closedform}A positive $\hat V$ is not yet evidence of event selection. In the empirical study of \S\ref{sec:setup}, the frozen rule is flat on $547$ of the events with genuine agent decisions (valid, non-fallback decisions) and the agent converts $294$ of them to long. If assigned at random, those $294$ interventions would each collect the average flat-event reaction of $+67$ bps (Table~\ref{tab:transdecomp})---a payoff that comes from the type of trade, not from which events were chosen. The benchmark makes this idea exact. Events are grouped into pools by the action the baseline took, optionally refined by a prespecified stratum $Z$; the primary specification uses no refinement. Year-stratified, firm-preserving, and firm-and-quarter demeaned variants are robustness checks, fixed in a post-registration analysis plan before any audit statistic was computed; \emph{prespecified} refers to that plan throughout. Within each pool, the observed number of each transition type is held fixed, and the events that receive those transitions are redrawn at random. Writing $\pi$ for such a reassignment and $\hat V^{\pi}$ for the deployment value recomputed under it, conditional on the event table and the observed counts,
\begin{equation}\label{eq:selection}
C_Z=\mathbb{E}_{\pi}\!\left[\hat V^{\pi}\right],\qquad
S_Z=\hat V-C_Z .
\end{equation}
The \emph{composition benchmark} $C_Z$ is the value the agent's transition composition earns without any event selection, and the \emph{selection value} $S_Z$ is what the agent's choice of events adds beyond it. Throughout, selection value means performance relative to the prespecified reassignment benchmark; where within-pool exchangeability is in doubt (\S\ref{sec:inference}), it should be read as a benchmark comparison rather than a causal effect.

The center $C_Z$ needs no simulation---it follows directly from the transition counts and each pool's average payoff; Monte Carlo draws serve only the reference distribution. A zero-centered test is correct only in the special case where these count-weighted pool averages cancel.

\subsection{Inference and scope}\label{sec:inference}\label{sec:uncertainty}
If the agent has no selection skill---its interventions land on random eligible events---the reassignment distribution is centered exactly at $C_Z$: $S_Z$ is centered at zero, while $\hat V$ generally is not. For the win statistic (the share of interventions with positive paired payoff), the no-skill center is likewise the pools' base rate, not necessarily $0.5$. Larger samples make this problem worse, not better: uncertainty around $\hat V$ shrinks while its center stays at $C_Z$, so a test centered at zero eventually flags almost every no-skill agent as skilled. The failure is a wrong reference point, not a wrong standard error; Table~\ref{tab:benchmark} quantifies it at this sample size ($.116$ vs.\ $.053$), and Figure~\ref{fig:stressgrid} traces it across a structural stress grid.

The randomization test rests on three conditions: (i)~within each pool, given the transition counts, a no-skill agent is equally likely to place its changes on any eligible events (\emph{exchangeability})---the substantive assumption, which the audit itself cannot establish; (ii)~the agent's own trades do not move returns, so every candidate action can be scored on the same realized return; and (iii)~each randomized cell contains both changed and unchanged eligible events, with boundary cells reported and excluded. When these hold, re-randomization reproduces the reference distribution exactly, and the $S_Z$ test requires no large-sample approximation \citep{fisher1935design,rosenbaum2002observational}.

Exchangeability is needed only to read the $p$-value as a formal test; the decomposition itself does not depend on it---$C_Z$ remains a transparent benchmark, and $S_Z$ a descriptive contrast, even where the assumption is contested. Section~\ref{sec:poscontrol} stress-tests it with no-skill agents whose interventions depend on event features, and results should be reported across conditioning sets chosen \emph{ex ante} (\S\ref{sec:results-diag}). Pools comprise genuine agent decisions only: events where the agent fails to produce a valid decision fall back to the prespecified default and contribute no interventions.

Uncertainty in $\hat V$ is quantified with a firm-cluster bootstrap ($44$ clusters in the application): treating events as independent gives intervals that are too narrow \citep{cameron2015practitioner}, and firm clustering is the most conservative prespecified choice (\S\ref{sec:results-diag}). When a decision time overlaps the return window, the primary audit scores on future-only returns (\S\ref{sec:results-ext}). One boundary is worth stating plainly: because the audit holds the number and ordered types of transitions fixed, skill in \emph{how often to intervene} and \emph{which transition types} to use accrues to $C_Z$; \emph{which eligible events} receive those transitions is what $S_Z$ measures. The year-stratified null in Table~\ref{tab:nullspec} additionally conditions on coarse calendar timing.
\section{Empirical Setting and Frozen Experiment}\label{sec:setup}
\subsection{Design status and timeline}\label{sec:design}
Policy arms, action mappings, scoring rules, and confirmatory comparisons were preregistered on the Open Science Framework before execution, with hash-frozen prompts and a versioned universe file (the registration identifier is withheld for double-blind review). The scaffold-free and matched-information cells were registered as amendments before execution; remaining analyses are exploratory or post-registration, labeled in the deviation log. The policy-value decomposition and the transition-matched audit were developed after all traces were frozen, so we treat the empirical analysis as \emph{retrospective} rather than confirmatory. Table~\ref{tab:flow} gives the sample flow and evaluation windows.

\begin{table}[t]
\centering
\caption{\textbf{Sample flow, evaluation windows, and decision timing.}}
\label{tab:flow}
\footnotesize
\setlength{\tabcolsep}{2.4pt}
\renewcommand{\arraystretch}{1.00}
\begin{tabular}{@{}lr l@{}}
\toprule
Stage & $N$ & Note \\
\midrule
Raw earnings events & $1{,}291$ & $1{,}219$ main $+$ $72$ megacap \\
Main-cohort candidate events & $917$ & 2020--2025 calendar window\\
Scoreable out-of-sample events & $723$ & CRSP availability; eff.\ $\le$2024 \\
\quad Full fallback (ITT default; counted) & $53$ & retries exhausted \\
Genuine agent decisions (per-protocol) & $670$ & $=723-53$ \\
Rule-overrides (agent vs.\ rule) & $310$ & $302$ flat$\to$trade $+\,8$ other \\
Scaffold-free arm eligible / genuine & $719/710$ & $4$ lack predictions; $9$ fallbacks \\
Matched-info.\ factorial cell scored & $719$ & initial API gap; one logged rerun \\
External $[0,+1]$ rescore & $719$ & $4$ lack the $[0,+1]$ leg \\
\midrule
\multicolumn{3}{@{}l}{\emph{Windows.} 2017--19 training; 2020--24 main (2024 = post-cutoff, $166$); 2025 unused.}\\
\multicolumn{3}{@{}l}{\emph{Handling.} ITT: intention-to-treat (fallbacks keep default); PP: per-protocol.}\\
\multicolumn{3}{@{}l}{\emph{Timing.} In-house inputs freeze at close of $t{-}2$ (no overlap);}\\
\multicolumn{3}{@{}l}{external framework decides at $t{-}1$ $\Rightarrow$ future-only $[0,+1]$ (\S\ref{sec:results-ext}).}\\
\bottomrule
\end{tabular}
\end{table}

\subsection{Data and policies}\label{sec:data}
Firms enter the universe under four frozen criteria---consumer-facing GICS classification (consumer discretionary or staples with a brand-driven or direct-to-consumer channel), NYSE/Nasdaq listing, coverage by at least three sell-side analysts at entry, and a measurable social footprint at entry. The resulting $44$-firm main cohort ($36$ discretionary, $8$ staples) was committed, with per-firm entry dates ($2017$-$01$-$01$, or IPO date if later) and identifiers, before any outcome data were scored. Two megacaps (Coca-Cola and Nike; KO/NKE) were logged as a separate megacap cohort from the outset (Table~\ref{tab:flow}) and are outside the committed $44$-firm universe, so they are excluded from the audited cohort. Events are quarterly earnings announcements. The forecaster's out-of-sample period is 2020--2024; the frozen CRSP extract ends in 2024. Separately, only the $166$ calendar-2024 events postdate the in-house backbone's knowledge cutoff (Dec-2023), forming the backbone-post-cutoff subset used in \S\ref{sec:results-diag}. Ground truth is the three-day Carhart four-factor $\car$ \citep{famafrench1993,carhart1997}, frozen in advance: loadings are estimated by OLS on days $[t{-}250,t{-}30]$ ($100$-observation minimum), and events failing estimation or return-window requirements are excluded in Table~\ref{tab:flow}.

Prices come from CRSP, accounting data from Compustat, and consensus forecasts from I/B/E/S with a $t{-}2$ cutoff. Social inputs come from timestamp-filtered X and Reddit posts linked through frozen firm-level queries (X capped at the $1{,}500$ most-recent matching posts per event; Reddit uncapped), with a point-in-time window ending at $t{-}2$. Google Trends provides retrospective search-volume features; \S\ref{sec:results-diag} therefore reports a no-Trends rerun.

The headline comparison is the \emph{full agent} (arm A0) versus the \emph{frozen-rule baseline} (arm A1, the registered comparator). The full agent runs three LLM roles over the deterministic forecaster: the \emph{Scout} filters raw social text into firm-relevant notes, the \emph{Strategist} reads those notes together with the structured forecast and its rule-implied default position and chooses the final position outright (it may accept the default or override it), and the \emph{Reviewer} validates the output before it is logged. The \emph{scaffold-free} arm gives a single LLM the same inputs without this division of roles. Additional system ablations remove the Scout (arm A2) or the social channel; \S\ref{sec:results-diag} reports both. A separate matched-information factorial crosses policy class (deterministic vs.\ LLM) with information set (structured-only vs.\ five prespecified social aggregates: three Trends as-of features, log X-post count, log X engagement); Table~\ref{tab:arms} defines its arms. A-series arms are complete systems; R/L-series arms are the re-fit factorial variants of Table~\ref{tab:arms}, not new systems. The full agent (A0) is scored against the registered frozen-rule baseline. Payoff is the primary outcome measure; win rates are descriptive. Fallbacks ($53/723$) are handled as intention-to-treat (ITT) and per-protocol (PP). A single frozen threshold rule converts each forecast into a default position: long when the forecast exceeds one forecast standard deviation ($0.00995$), short below the negative threshold, flat otherwise. Deterministic arms trade this rule position as is; in LLM arms it enters the prompt only as the default, so any disagreement between an LLM arm and the baseline is an explicit override logged in the decision trace. In-house agents run Llama-3.3-70B-Instruct \citep{grattafiori2025llama,llama33card} with $n{=}3$ samples, temperature $0.3$, one logged retry, and hash-frozen prompts; each decision's final action is the majority vote of the three sampled outputs (the Scout instead takes the union of its samples' escalations). Provider-reported cutoffs, as of the July 2026 runs, are Dec-2023 for Llama and 2025-12-01 for GPT-5.5 \citep{openai2026gpt55}. GPT-5.5 is therefore a contamination-exposed positive control, not external validation  (\S\ref{sec:results-ext}). The deterministic substrate is an elastic net \citep{zou2005elasticnet} over 17 look-ahead-audited pre-event features (cross-validated information coefficient, IC, $+0.047$; out-of-sample IC $+0.006$); the headline comparison therefore measures what the agent layer adds to a weak but leakage-audited forecaster. The external framework (TradingAgents v0.2.5\footnote{Release v0.2.5, \url{https://github.com/TauricResearch/TradingAgents}; commit identifiers archived in the deviation log.} \citep{tradingagents}) maps forecasts to positions by its prespecified conservative rule and serves as the external stress test of \S\ref{sec:results-ext}.

\begin{table}[t]
\centering
\caption{\textbf{Matched-information factorial cells.} Prompts are hash-frozen.}
\label{tab:arms}
\footnotesize
\setlength{\tabcolsep}{3.2pt}
\renewcommand{\arraystretch}{1.00}
\begin{tabular}{@{}llll@{}}
\toprule
Arm & Policy class & Information set & Role \\
\midrule
$R_0$ & Deterministic & Structured only & Factorial control \\
$R_1$ & Deterministic & $+$ five aggregates & Factorial treated \\
$L_0$ & LLM & Structured only & Factorial control \\
$L_1^{\mathrm{agg}}$ & LLM & $+$ five aggregates & Matched-information cell \\
\bottomrule
\end{tabular}
\end{table}

Net value compares the two policies' standalone net returns, charging the pre-stated $40$ bps to each nonzero event position. This is a policy-level cost ($296$ net additional positions; $16.4$ bps/event induced cost; $37.1$ bps break-even per net additional position), not the literal cost of switching an executed position. The prespecified $40$ bps is a conservative round-trip friction; gross deployment value is the primary quantity, with $20$/$40$/$60$ bps overlays as sensitivities $+6.6$/$-2.1$/$-10.7$ bps/event under the switching-cost convention, which charges $c\,\lvert a^{A}_e-a^{B}_e\rvert$ per event. The primary economic test is one-sided (net value at most zero); the selection test is two-sided. Intervals are percentile firm-cluster bootstrap intervals with $10^4$ draws, and the randomization test uses $5{,}000$ draws with the standard add-one Monte Carlo $p$-value; two-sided values double the smaller tail with ties included.

\section{Evaluator Validation}\label{sec:poscontrol}
The rejection rates in this section describe the evaluator itself, not the real agent. The validation asks three questions. First, does the audit recover selection skill when skill is deliberately injected? Second, does transition matching prevent passive exposure from being attributed to selection? Third, does future-only scoring remove artificial signal created by decision-time overlap? §5.2 addresses  the first; §5.3 the second and third; §5.4 stress-tests the exchangeability assumption.

\subsection{Benchmark construction}\label{sec:benchconstruct}
Using the frozen event table and transition composition of \S\ref{sec:setup}, we construct semi-synthetic agents: the returns, eligible baseline-action pools, and transition counts are the frozen experiment's, and only the assignment of transitions to events is synthetic. The benchmark contains four scenarios. \emph{Centered null} uses sign-symmetrized returns and no selection. \emph{Exposure confound} restores the realized return distribution without selection. \emph{Known selection} tilts flat$\to$long assignments toward higher-payoff events ($q=0.60$). \emph{Timing overlap} assigns interventions using a pre-decision-observable return leg while scoring an overlapping window. Each replication holds the realized transition counts fixed ($294$ flat$\to$long and $16$ other changes) and assigns them to events without replacement from the eligible pools; under known selection, a Binomial($294$, $q$) number of the flat$\to$long picks are drawn uniformly from positive-reaction flat-pool events and the remaining picks from non-positive-reaction flat-pool events; all other transition types are drawn uniformly from their eligible pools. Payoffs are always the real recorded returns.

\subsection{Recovery of known selection skill}\label{sec:recovery}
\begin{figure}[t]
\centering
\includegraphics[width=\columnwidth]{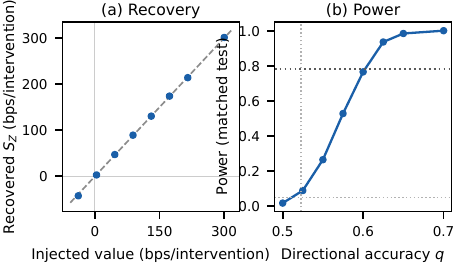}
\Description{Two side-by-side panels: recovered selection value against injected value along the identity line, and the power curve of the matched test rising with directional accuracy, matching 0.783 at q equal to 0.60.}
\caption{\textbf{Known-skill calibration.} Semi-synthetic agents with directional accuracy $q$ on their flat$\to$long picks, real returns, and the real transition composition ($R{=}2{,}000$ replications per point, $600$ reassignment draws each). \emph{(a)} Recovered vs.\ injected selection value (dashed: identity). \emph{(b)} Power of the one-sided matched test (dotted: Table~\ref{tab:benchmark}'s $.783$ at $q{=}0.60$; vertical guide: pool base accuracy $0.523$).}
\label{fig:calibration}
\end{figure}

We first verify that the evaluator recovers known selection skill: synthetic agents whose flat$\to$long picks have a known directional accuracy $q$, matched to the real agent's intervention rate and transition composition, pass through the unmodified pipeline; Figure~\ref{fig:calibration} reports the calibration from a replication of the generator. The mean recovered selection value tracks the injected value along the identity line, crossing zero not at $q=0.5$ but at the pool's base accuracy of $0.523$---the share of eligible flat-pool events with a positive reaction. Power of the one-sided matched test reaches Table~\ref{tab:benchmark}'s $.783$ at $q=0.60$ within Monte Carlo error, and the replication reproduces the matched audit's exposure cell ($.049$ vs.\ $.053$).

\subsection{Evaluator-component benchmark}\label{sec:components}
We compare seven procedures, $5{,}000$ replications per cell, using one-sided tests at $\alpha=.05$ ($R{=}1{,}000$/$R{=}800$ for the covariate and grid designs; the application's selection test, by contrast, is two-sided). The procedures successively add event pairing, fixed intervention counts, fixed ordered transition counts, and future-only scoring; Table~\ref{tab:benchmark} marks pairing, ordered-transition matching, and future-only scoring as P, K, and F. The intervention-count permutation (P without K) preserves only the total number of changes, while the transition-matched audit also preserves the ordered transition types.

\begin{table}[t]
\centering
\caption{\textbf{Evaluator-component benchmark.} One-sided false-attribution rates under three no-skill designs, and power under known selection ($q{=}0.60$).}
\label{tab:benchmark}
\footnotesize
\setlength{\tabcolsep}{2.0pt}
\renewcommand{\arraystretch}{1.02}
\begin{tabular}{@{}lccc rrr r@{}}
\toprule
 & & & & \multicolumn{3}{c}{False attribution (no skill)} & Power \\
\cmidrule(lr){5-7}\cmidrule(l){8-8}
Procedure & P & K & F & Centered & Exposure & Timing & $q{=}0.60$ \\
\midrule
Standalone book, i.i.d. & -- & -- & -- & .050 & .139 & .578 & .890 \\
Standalone book, clustered & -- & -- & -- & .053 & .200 & .577 & .917 \\
Unpaired ablation & -- & -- & -- & .010 & .014 & .164 & .625 \\
Zero-centered paired test & $\checkmark$ & -- & -- & .057 & .116 & .592 & .846 \\
Intervention-count permutation & $\checkmark$ & -- & -- & .048 & .021 & .490 & .665 \\
Transition-matched audit & $\checkmark$ & $\checkmark$ & -- & .052 & .053 & .648 & .783 \\
\textbf{Audit $+$ future-only scoring} & $\checkmark$ & $\checkmark$ & $\checkmark$ & \textbf{.052} & \textbf{.053} & \textbf{.000} & \textbf{.783} \\
\bottomrule
\end{tabular}

\smallskip
\begin{minipage}{\columnwidth}\footnotesize\emph{Note:} $5{,}000$ replications; Monte Carlo half-widths $\le0.014$. Standalone book: the agent's own return series tested against zero; unpaired ablation: reassignment without event pairing. P adds pairing, K conditions on the realized (ordered) transition composition, F uses future-only scoring under decision-time overlap.\end{minipage}
\end{table}

Composition matching corrects exposure misattribution. Under a centered null, most procedures are approximately correctly sized, so the exposure column reflects the choice of benchmark rather than miscalibration. Pairing (P) alone fixes neither error, and clustering a standalone (unpaired) return series increases false attribution ($.139\to.200$). Conditioning on the realized composition (K) reduces exposure false attribution
from $11.6\%$ to $5.3\%$ ($17.1\%$ when the paired test is firm-clustered; Table~\ref{tab:covdgp}), and future-only scoring (F) removes the timing artifact ($.648\to.000$). Correct centering costs some power ($.783$ vs.\ $.846$--$.917$ for the over-rejecting procedures).

Timing overlap requires a separate correction. Composition matching does not remove decision-time overlap---it concentrates overlap in the flat$\to$long cell ($.592\to.648$)---so future-only scoring is a mandatory companion. Because the overlap design places the entire artificial signal in the pre-decision leg, the $.000$ entry is a signal-removal check; under signal-free designs, the audit's false-alarm rate is the centered-null column (.052). 

\subsection{Stress tests and out-of-period replication}\label{sec:stress}
\begin{table}[t]
\centering
\caption{\textbf{Size and exchangeability stress tests.}
Rejection rates under clean no-skill assignments (U, SX),
covariate-dependent assignments (RX, NL), and injected power (PW)
($R{=}1{,}000$; one-sided $5\%$; real returns and transition counts).}
\label{tab:covdgp}
\footnotesize
\setlength{\tabcolsep}{2.3pt}
\renewcommand{\arraystretch}{1.00}
\begin{tabular}{@{}lrrrrr@{}}
\toprule
& \multicolumn{4}{c}{Rejection / sensitivity rate} & Power \\
Assignment / test & U & SX & RX & NL & PW \\
\midrule
\textbf{Matched, baseline pools (primary)} & $.050$ & $.061$ & $.013$ & $.020$ & $.643$ \\
Matched, $\times$ year & $.045$ & $.060$ & $.025$ & $.046$ & $.702$ \\
Matched, $\times$ firm & $.057$ & $.069$ & $.040$ & $.022$ & $.770$ \\
Zero-centered paired, clustered & $.171$ & $.173$ & $.066$ & $.107$ & $.861$ \\
\bottomrule
\end{tabular}

\smallskip
\begin{minipage}{\columnwidth}\footnotesize\emph{Note:} Intervention probabilities: uniform (U); proportional to a synthetic covariate independent of returns (SX); proportional to $\exp(0.8\,\mathrm{social}_z{+}0.8\,\mathrm{vol}_z{-}0.4\,\mathrm{size}_z)$ over three real pre-decision covariates---one-month volatility, log assets, 30-day X post count---that also shift the return distribution (RX); from a misspecified nonlinear model adding firm and year effects, a volatility$\times$social interaction, and a threshold (NL). 
U and SX are clean no-skill size designs. Because RX and NL use
return-shifting pre-decision features, their rejection rates diagnose sensitivity to exchangeability misspecification rather than formal Type-I error.
PW: power under an injected tilt of $\approx\!+115$ bps/intervention. Monte Carlo half-widths $\le\!1.6$\,pp (matched rows), $\le\!2.4$\,pp (zero-centered), $\le\!3.0$\,pp (power).\end{minipage}
\end{table}

A real agent may be more likely to intervene on some kinds of events than others, which is exactly how exchangeability could fail. Table~\ref{tab:covdgp} contrasts clean no-skill size designs (U/SX) with covariate-dependent sensitivity designs (RX/NL). U/SX stay near nominal. When return-shifting features drive assignment, rejection falls to $.013$--$.020$ on this event table; these RX/NL rates are sensitivity diagnostics, not formal false-attribution rates. Under NL, up to $7.5\%$ of interventions lack eligible alternatives and are excluded. Richer conditioning raises injected-tilt power ($.770$ vs.\ $.643$), while the zero-centered paired test rejects more often in every design ($.066$--$.173$).

\begin{figure*}[t]
\centering
\includegraphics[width=0.70\textwidth]{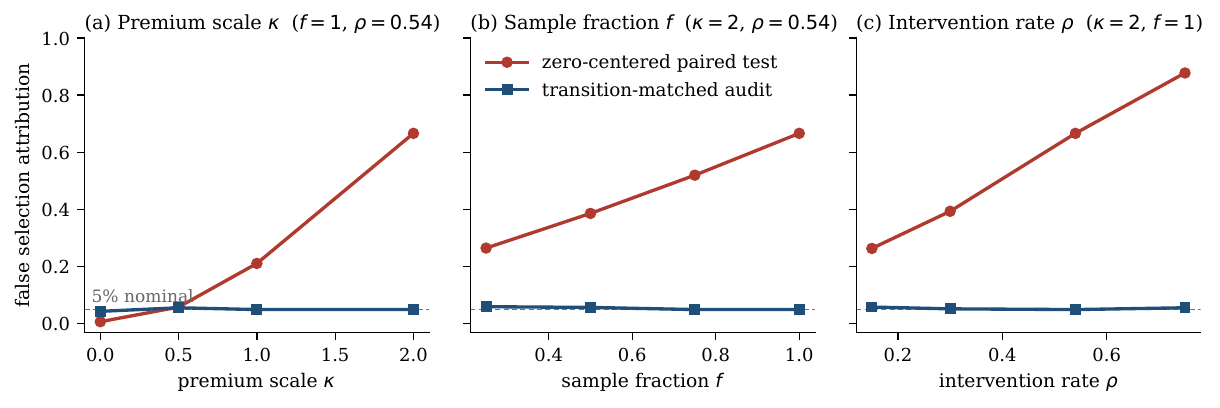}
\Description{Three line panels: false-attribution rates rise toward 0.88 for the zero-centered test and stay near 0.05 for the matched audit as premium scale, sample fraction, and intervention rate increase.}
\caption{\textbf{Structural stress grid.} One-sided false-attribution rates for no-skill flat$\to$long agents across premium scale $\kappa$, sample fraction $f$, and intervention rate $\rho$ (slices through the $64$-cell grid; $R{=}800$ per cell; dashed line: $5\%$ nominal).}
\label{fig:stressgrid}
\end{figure*}

The structural stress grid extends this beyond the headline cell. As the premium scale $\kappa$, sample fraction $f$, and intervention rate $\rho$ grow, the zero-centered test drifts further from the correct reference point---its false-attribution rate reaches $.88$---while the audit stays near nominal in every cell ($.04$--$.07$; Figure~\ref{fig:stressgrid}). At every scale, the failure is the wrong center, not the wrong standard error (\S\ref{sec:inference}).

Finally, the exposure and known-selection cells were replicated out of period, on the $285$ pre-2020 main-cohort events with valid outcomes. These events were used to train the forecaster but never entered any agent evaluation, so this is an out-of-period evaluator replication, not a held-out model test. The design is a simplified single pool (flat baseline; $153$ flat$\to$long interventions, matching the audit's flat-pool intervention share). Zero-centered false attribution is $8.4\%$ against $4.9\%$ for the matched audit; matched-test power at $q{=}0.60$ is $.65$ on this smaller table ($R{=}5{,}000$ and $2{,}000$; half-widths $\le0.9$ and $2.1$\,pp); and the pre-2020 pool's base accuracy ($0.523$) and mean reaction ($+69$ bps) closely match the audited pool's.

\section{Empirical Agent Audit}\label{sec:results}

\subsection{Deployment value}\label{sec:results-primary}
The agent's gross deployment value is positive but imprecisely estimated, and approximately zero net of the prespecified cost. The agent changes $310$ of $723$ baseline actions, and the gross estimate is $+15.2$ bps/event, with a firm-clustered $95\%$ CI of $[-45.2,+69.2]$; per-protocol agrees ($+16.4$). The net estimate is $-1.2$ bps/event, with a $95\%$ CI of $[-59.8,+54.3]$, and the one-sided test that net value is at most zero does not reject ($p=0.52$). Table~\ref{tab:policyvalue} reports the main policy-value contrasts (factorial: Table~\ref{tab:factorial}).

\subsection{Selection attribution}\label{sec:results-selection}
The positive deployment value is explained by the agent's transition composition rather than by a detectable event-selection advantage. Holding the observed transition composition fixed ($294$ flat$\to$long, $8$ flat$\to$short, $8$ other) gives a matched mean of $60.1$ bps per intervention against the agent's observed $35.5$. The agent lies at the $32$nd percentile of the randomization distribution ($p=0.63$); the selection value is $-10.6$ bps/event ($-24.6$ per intervention). The win rate agrees: $0.500$ lies at the $13$th percentile of the composition-correct null, whose base rate is $0.521$, not $0.5$ ($p=0.33$); the registered symmetric-$0.5$ null understates this base rate (\S\ref{sec:inference}).

\begin{figure}[t]
\centering
\includegraphics[width=\columnwidth]{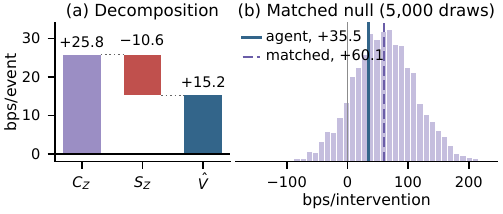}
\Description{Two panels: a bar decomposition of deployment value into composition benchmark and selection value, and a histogram of the matched reassignment distribution with the observed value at its 32nd percentile.}
\caption{\textbf{Audit of the observed interventions.} \emph{(a)} Event-level decomposition of $\hat V$ into $C_Z$ and $S_Z$. \emph{(b)} The agent's payoff per intervention within the matched reassignment distribution (Table~\ref{tab:transdecomp}).}
\label{fig:forest}
\end{figure}

At the event level, the observed $+15.2$ bps decomposes into a $+25.8$ composition-benchmark component and a $-10.6$ selection component (Figure~\ref{fig:forest}; Table~\ref{tab:transdecomp}). The design has $80\%$ power only at ${\approx}142$ bps/intervention (\S\ref{sec:results-diag}), so this null result is informative against large selection skill but not against small effects. The result supports a statement about attribution rather than equivalence: the deployment interval ($[-45.2,+69.2]$ bps/event) remains too wide to rule out economically meaningful deployment effects in either direction. Table~\ref{tab:transdecomp} reports the decomposition by transition type: the gross increment comes from the $294$ flat$\to$long interventions, whose realized payoffs are below those of matched random interventions, while the sparse short and reversal legs ($16$ events) are noisy, with a net $-4.2$ bps/event drag.

\begin{table}[t]
\centering
\caption{\textbf{Deployment value} (bps/event; gross, and net of the pre-stated $40$ bps cost). Firm-clustered $95\%$ CIs on the gross estimate ($B_{\mathrm{boot}}=10^4$); quarter-clustered confidence intervals are similar. External row: TradingAgents with the Llama backbone, future-only $[0,+1]$ window; ``social $-$ orig'' compares its social-augmented and original configurations (\S\ref{sec:results-ext}).}
\label{tab:policyvalue}
\footnotesize
\setlength{\tabcolsep}{2.6pt}
\renewcommand{\arraystretch}{1.00}
\begin{tabular}{@{}lrrrrr@{}}
\toprule
Contrast & $N$ & Chg.\ rate & Gross & Net & $95\%$ CI \\
\midrule
\textbf{Full agent vs.\ rule (ITT)} & $723$ & $0.43$ & $\mathbf{+15.2}$ & $\mathbf{-1.2}$ & $[-45.2,+69.2]$ \\
Scaffold-free LLM vs.\ rule (PP) & $710$ & $0.71$ & $-7.0$ & $-28.4$ & $[-138.2,+101.7]$ \\
External Llama: social $-$ orig & $719$ & $0.45$ & $-18.5$ & --- & $[-100.1,+56.1]$ \\
\bottomrule
\end{tabular}
\end{table}

\begin{table}[t]
\centering
\caption{\textbf{Transition-level decomposition} (bps/intervention except the last column). Per-transition null means are exact pool means; their intervention-weighted average is the exact $+60.1$ (Monte Carlo mean $+59.5$). ``Other'' $=$ $4$ long$\to$flat, $2$ short$\to$flat, $2$ long$\to$short.}
\label{tab:transdecomp}
\footnotesize
\setlength{\tabcolsep}{3.2pt}
\renewcommand{\arraystretch}{1.00}
\begin{tabular}{@{}lrrrrr@{}}
\toprule
Transition & $n$ & Observed & Null mean & Residual & To $\hat V$ (bps/event) \\
\midrule
flat$\to$long & $294$ & $+47.7$ & $+67.4$ & $-19.7$ & $+19.4$ \\
flat$\to$short & $8$ & $+145.1$ & $-67.4$ & $+212.5$ & $+1.6$ \\
other & $8$ & $-522.8$ & $-79.4$ & $-443.4$ & $-5.8$ \\
\textbf{Total} & $\mathbf{310}$ & $\mathbf{+35.5}$ & $\mathbf{+60.1}$ & $\mathbf{-24.6}$ & $\mathbf{+15.2}$ \\
\bottomrule
\end{tabular}
\end{table}

\subsection{Integrity and robustness}\label{sec:results-diag}\label{sec:results-decomp}
The attribution conclusion is stable across the reported specifications, although the study remains underpowered for small effects. The factorial tests the information channel directly. Holding the five aggregate features fixed across policy classes reveals neither a detectable information effect nor an information$\times$policy interaction (Table~\ref{tab:factorial}): the re-fit deterministic policy barely reacts to the aggregates, and the two policy-class contrasts nearly coincide across feature sets. The aggregates change LLM behavior far more than value (the LLM still disagrees with $R_1$ on $59\%$ of events), with value consequences indistinguishable from zero. The audit reads the other in-house pairs the same way: the scaffold-free arm ($505$ changes, a different mix) splits into a $+25.1$ composition benchmark and $-34.9$ selection value per intervention ($20$th percentile, $p=0.41$); the Scout-removal arm (A2, $164$ changes) into $+32.9$ and $+16.2$ ($58$th, $p=0.85$). Trading style moves the mechanical component; no pair shows a detectable event-selection advantage.

The placebos give the behavioral counterpart: wrong-firm social collapses the intervention rate among genuine decisions from $0.463$ to $0.029$ and name-scrubbing partially restores it to $0.281$; although intervention rates shift sharply, all corresponding return-contrast confidence intervals include zero---entity alignment drives trading frequency without measurably changing trade value. Time- and outcome-shifted placebos leave the conclusion in place, as does removing the retrospective Trends series, which enters only the agent's social-feature block (the rule's actions are identical without it): gross deployment value $+15.4$ bps/event; per intervention, $C_Z=+48.9$ and $S_Z=-13.4$ ($40$th percentile, $p=0.80$).

\begin{table}[t]
\centering
\caption{\textbf{Matched-information factorial.} Gross bps/event with firm-clustered $95\%$ CIs; both policy classes receive the same five aggregate social features.}
\label{tab:factorial}
\footnotesize
\setlength{\tabcolsep}{3.6pt}
\renewcommand{\arraystretch}{1.00}
\begin{tabular}{@{}lrrr@{}}
\toprule
Contrast & Change rate & Gross & $95\%$ CI \\
\midrule
Social info., deterministic ($R_1{-}R_0$)$^{\ast}$ & $0.01$ & $+0.5$ & $[-7.5,+7.2]$ \\
Social info., LLM ($L_1^{\mathrm{agg}}{-}L_0$) & $0.41$ & $+1.3$ & $[-58.0,+64.6]$ \\
Policy, matched info.\ ($L_1^{\mathrm{agg}}{-}R_1$) & $0.59$ & $-31.9$ & $[-149.0,+86.8]$ \\
Policy, structured only ($L_0{-}R_0$) & $0.76$ & $-32.7$ & $[-171.5,+96.1]$ \\
Information$\times$policy interaction & --- & $+0.8$ & $[-57.6,+62.0]$ \\
\bottomrule
\end{tabular}

\smallskip
\begin{minipage}{\columnwidth}\footnotesize\emph{Note:} $719$-event common support (the four events lacking predictions remove one firm). $^{\ast}$Deterministic policies re-fit at a fixed prespecified regularization ($293/292$ trades). Dashes: not defined for derived contrasts.\end{minipage}
\end{table}

\begin{table}[t]
\centering
\caption{\textbf{Null-specification robustness.} Percentiles of the observed payoff/intervention and win rate under each transition-matched null ($5{,}000$ draws).}
\label{tab:nullspec}
\footnotesize
\setlength{\tabcolsep}{3.4pt}
\renewcommand{\arraystretch}{1.00}
\begin{tabular}{@{}llrr@{}}
\toprule
Null specification & Conditioning & Payoff pctl. & Win pctl. \\
\midrule
\textbf{Primary} & baseline action & $32$nd & $13$th \\
Year-stratified & $\times$ year & $34$th & $15$th \\
Firm-preserving$^{\ddagger}$ & within firm & $13$th & $12$th \\
Demeaned$^{\S}$ & firm $+$ quarter & $22$nd & --- \\
Propensity-stratified$^{\dagger}$ & $\times$ propensity quintile & $28$th & $12$th \\
\bottomrule
\end{tabular}

\smallskip
\begin{minipage}{\columnwidth}\footnotesize\emph{Note:} $^{\dagger}$Exploratory; other rows prespecified; every primary-specification cell retains eligible alternatives. $^{\ddagger}$Support requirement excludes five within-firm cells ($13$ interventions); included-support payoff $8.1$ bps/intervention (win $0.495$). $^{\S}$Payoffs demeaned by firm and quarter before reassignment; the win percentile is not defined for this statistic.\end{minipage}
\end{table}

The attribution conclusion is qualitatively unchanged across the reported conditioning, clustering, and timing specifications (Table~\ref{tab:nullspec}); the propensity-stratified row (pools split by quintiles of the estimated chance that the agent intervenes on an event) is backed by a diagnostic showing that every intervention has eligible alternatives in its cell; and alternative date, week, and two-way clusterings \citep{cameron2011robust} produce smaller standard errors than firm clustering (\S\ref{sec:inference}). The firm-preserving row also documents concentration: the $13$ interventions in its five excluded cells carry most of the gross intervention payoff (included-support mean $8.1$ vs.\ $35.5$ bps/intervention), so the positive deployment estimate rests on a few high-payoff events. The attribution conclusion does not: in all $44$ leave-one-firm-out reruns the selection value stays negative and remains at or below the matched-null median (percentiles $17$--$50$, $p\ge0.33$).

Decision-level evidence agrees: the registered win-rate test gives $155/310=0.500$ (McNemar's paired test concurs \citep{mcnemar1947}), interventions are stable across the three logged samples ($84.5\%$ unanimous, zero flips), and with the social channel removed the system reproduces the rule on $723/723$ events, so every action change is attributable to the social channel.

The post-cutoff subset and the power analysis bound what the null result can mean. On the $166$ calendar-2024 events past the in-house backbone's Dec-2023 cutoff, the agent changes $68$ actions for $+78.9$ bps/event gross ($[-38.6,+202.5]$; net $+62.5$). The audit reaches the same inferential conclusion, with the per-intervention point estimate moving from below to above the matched mean ($192.6$ vs.\ $134.5$, $68$th percentile)---consistent with sampling noise at this interval width, which remains too wide to exclude meaningful differences. Approximating a true effect as a constant shift of the observed distributions, the design reaches $80\%$ power at two-sided $5\%$ for a selection value of ${\approx}142$ bps/intervention ($61$ bps/event), and the firm-clustered deployment test at ${\approx}82$ bps/event.

\subsection{External stress test}\label{sec:results-ext}
On the identical universe, TradingAgents v0.2.5 with the Llama backbone produces an imprecise future-only social contrast, $-18.5$ $[-100.1,+56.1]$, and its post-cutoff 2024 subset agrees ($-68.9$ $[-225.7,+81.9]$). Across the in-house arms, no standalone mean return survives a Romano--Wolf stepdown over the arm family (one-sided FWER $5\%$, joint event bootstrap; smallest adjusted $p=0.31$) \citep{romanowolf2005}. GPT-5.5 is retained as a \emph{contamination-exposed positive control} because its provider-reported cutoff postdates the entire evaluation window (\S\ref{sec:data}); its ``social $-$ orig'' contrast, under the same framework and future-only window, is $+13.9$ $[-18.0,+45.8]$, and announcement-aligned rescoring (from $+16.8$ conventional to $+2.0$ on the rescoring subsample) indicates that part of this contrast reflects pre-release exposure.

\section{Conclusion}\label{sec:conclusion}
A paired return answers whether deploying an agent changes realized outcomes; it does not establish that the agent selected the right events. In controlled benchmarks, preserving the observed transition composition restores the appropriate no-selection reference, reducing false attribution from $11.6\%$ to $5.3\%$. In the empirical application, the audit separates a positive gross deployment estimate from a negative selection estimate that is statistically indistinguishable from zero. Reporting both quantities---deployment value and performance against the transition-composition benchmark---makes the attribution target explicit for event-aligned financial-agent evaluations.

\section{Discussion and Limitations}\label{sec:discussion}
Matching the realized transition composition changes the benchmark, not the standard error: the no-selection null moves from zero to the payoff of random interventions of the same types (Table~\ref{tab:benchmark}). In the application, this re-centering separates the positive gross deployment estimate from the negative selection estimate, and the factorial and placebos trace the agent's frequent interventions to entity alignment rather than measurable information value. A positive deployment value with little selection value may still justify deployment net of cost but does not support a claim of event-selection skill; conversely, an agent can show selection skill yet fail to justify deployment when its interventions are too sparse or costly.

The audit applies when two policies can be scored on the same events, payoffs are observable for each candidate action, the audited actions do not materially affect realized outcomes, and credible reassignment pools can be specified. It is not a substitute for off-policy evaluation when rewards for unchosen actions are unobserved, nor does it directly cover sequential policies whose actions alter later states.

The selection result remains conditional on $Z$ and on exchangeability, which the audit assumes rather than verifies; the feature-dependent simulations probe it only on this event table. The procedure evaluates final policy actions rather than internal reasoning. The application is retrospective and lacks a held-out 2025 confirmation; the factorial is complete only at the aggregate representation, and residual contamination is not eliminated (the $166$-event post-cutoff subset reproduces the conclusion but is underpowered; \S\ref{sec:results-diag}). The deployment interval is wide, the application covers one event class at 44 firms, Google Trends is retrospective, and event-time alignment is at daily resolution, so the conclusions are setting-specific. Social inputs are platform- and query-selected (including the $1{,}500$-post X cap), $53$ of $723$ events use fallback defaults, and the long/short/flat action map omits position sizing, further limiting external validity.

The results do not establish that the agent has zero value, nor do they validate the exchangeability assumption; selection value remains conditional on the reported benchmark and conditioning set. Leakage diagnostics remain a separate and necessary integrity check, because a leaked feature invalidates every return-based procedure, including ours.

\section*{Reproducibility, Ethics, and Data}
All reported tables and figures are reproducible from de-identified intermediate data; upon acceptance, we will release the benchmark generator, analysis code, frozen prompt hashes, the OSF preregistration record, deviation log, and per-decision traces, subject to data licenses (WRDS inputs not redistributed); released artifacts aggregate public posts per event and exclude user identifiers and post-level text. This is a scientific evaluation, not investment advice.

\bibliographystyle{ACM-Reference-Format}
\bibliography{references_cs}

\end{document}